\def\year{2019}\relax
\documentclass[letterpaper]{article} 
\usepackage{aaai19}  
\usepackage{times}  
\usepackage{helvet}  
\usepackage{courier}  
\usepackage{url}  
\usepackage{graphicx}  
\usepackage{multirow, booktabs}
\usepackage{amsmath}
\usepackage{amssymb}
\usepackage{amsfonts}
\usepackage{makecell}

\usepackage{verbatim}
\usepackage{url}
\usepackage{array}
\usepackage{xcolor}
\usepackage{soul}
\newcolumntype{L}[1]{>{\raggedright\let\newline\\\arraybackslash\hspace{0pt}}m{#1}}
\setul{2pt}{2pt}

\usepackage{pifont}
\newcommand{\xmark}{\ding{55}}

\begin{document}
%
\title{A Unified Model for Opinion Target Extraction and Target Sentiment Prediction\thanks{The work described in this paper was mainly done when Lidong Bing was an employee and Xin Li was an intern at Tencent AI Lab. It is substantially supported by grants from the Research Grant Council of the Hong Kong Special Administrative Region, China (Project Codes: 14203414) and the Direct Grant of the Faculty of Engineering, CUHK (Project Code: 4055093).}}
\author{
Xin Li,\textsuperscript{\rm 1}
Lidong Bing,\textsuperscript{\rm 2}
Piji Li,\textsuperscript{\rm 3}
Wai Lam\textsuperscript{\rm 1}\\
\textsuperscript{\rm 1}Department of Systems Engineering and Engineering Management,\\ The Chinese University of Hong Kong\\
\textsuperscript{\rm 2}R\&D Center Singapore, Machine Intelligence Technology, \\ Alibaba DAMO Academy\\
\textsuperscript{\rm 3}Tencent AI Lab, Shenzhen, China \\
\{lixin, wlam\}@se.cuhk.edu.hk, l.bing@alibaba-inc.com, pijili@tencent.com
}
\maketitle
\begin{abstract}
Target-based sentiment analysis involves opinion target extraction and target sentiment classification. However, most of the existing works usually studied one of these two sub-tasks alone, which hinders their practical use. This paper aims to solve the complete task of target-based sentiment analysis in an end-to-end fashion, and presents a novel unified model which applies a unified tagging scheme. Our framework involves two stacked recurrent neural networks: The upper one predicts the unified tags to produce the final output results of the primary target-based sentiment analysis; The lower one performs an auxiliary target boundary prediction aiming at guiding the upper network to improve the performance of the primary task. To explore the inter-task dependency, we propose to explicitly model the constrained transitions from target boundaries to target sentiment polarities. We also propose to maintain the sentiment consistency within an opinion target via a gate mechanism which models the relation between the features for the current word and the previous word. We conduct extensive experiments on three benchmark datasets and our framework achieves consistently superior results.
\end{abstract}

\section{Introduction}
Target-Based Sentiment Analysis (TBSA) aims to detect the opinion targets explicitly mentioned in sentences and predict the sentiment polarities over the opinion targets~\cite{liu2012sentiment,S14-2004}. For example, in the sentence ``\textbf{\textit{USB3 Peripherals}} \textit{are noticably less expensive than the} \textbf{\textit{ThunderBolt ones}}'', the user mentions two opinion targets, namely, ``\textit{\textbf{USB3 Peripherals}}'' and ``\textit{\textbf{ThunderBolt ones}}'', and expresses positive sentiment over the first, and negative sentiment over the second. 

Traditionally, this task can be broken into two sub-tasks, namely, opinion target extraction and target sentiment classification. The goal of opinion target extraction is to detect the opinion target mentions in the text, and it has been extensively studied~\cite{J11-1002,P13-1172,P14-1030,D15-1168,yin2016unsupervised,D16-1059,wang2017coupled,P17-1036,D17-1310,li2018aspect,P18-2094}. 
The second sub-task, i.e., target sentiment classification, performs as a multiplier for the usefulness of the extracted target mentions, as it can predict the sentiment polarity of the given opinion targets.
This sub-task has also received a lot of attention in recent years~\cite{P14-2009,D16-1021,D16-1058,ma2017interactive,D17-1047,tay2017learning,ma2018targeted,N18-2043,P18-1087,P18-1088,P18-1234,P18-2092,li2019exploiting}. However, most existing methods solving the second sub-task assume that the target mentions are given, which limits their practical use. To sum up, all the above works aim at solving only one of the sub-tasks. In order to apply these existing methods in practical settings, i.e., not only extracting the targets, but also predicting the target sentiment, one typical way is to pipeline the methods of the two sub-tasks together. 

As observed in some other tasks~\cite{W03-1026,W04-3236,N09-1037,D14-1200}, if two sub-tasks have strong couplings (e.g, NER and relation extraction), a more integrated model is usually more effective than a pipline solution. For the TBSA task, previous researchers have attempted two approaches to a more integrated solution~\cite{D13-1171,D15-1073}. One approach is to make the models of the two sub-tasks jointly trained, which utilizes a set of target boundary tags (e.g., \texttt{B}, \texttt{I}, \texttt{E}, \texttt{S} and \texttt{O}) and a set of sentiment tags (e.g. \texttt{POS}, \texttt{NEG}, \texttt{NEU}). The ``joint'' row of Table~\ref{tab:tagging} gives an example of the tagging scheme in this approach. Another approach is to totally dismiss the boundary of the two sub-tasks, which utilizes a set of specially-designed tags (we name it ``unified tagging scheme''), namely, \texttt{B-\{POS, NEG, NEU\}}, \texttt{I-\{POS, NEG, NEU\}}, \texttt{E-\{POS, NEG, NEU\}}, \texttt{S-\{POS, NEG, NEU\}}, denoting the beginning of, inside of, end of, and single-word opinion target with positive, negative or neutral sentiment respectively, and \texttt{O} denoting NULL sentiment. An example is given in the ``unified'' row in Table~\ref{tab:tagging}. Unfortunately, these initial attempts did not result in a more integrated model that can outperform the pipeline approaches.

\begin{table*}[!t]
    \centering
    \begin{tabular}{lccccccccccccc}
    \Xhline{2\arrayrulewidth}
        Input & The & AMD & Turin & Processor & seems & to & always & perform & much & better & than & Intel & . \\ \hline
        \multirow{2}{*}{Joint}  & \texttt{O} & \texttt{B} & \texttt{I} & \texttt{E} & \texttt{O} & \texttt{O} & \texttt{O} & \texttt{O} & \texttt{O} & \texttt{O} & \texttt{O} & \texttt{S} & \texttt{O}  \\
         & \texttt{O} & \texttt{POS} & \texttt{POS} & \texttt{POS} & \texttt{O} & \texttt{O} & \texttt{O} & \texttt{O} & \texttt{O} & \texttt{O} & \texttt{O} & \texttt{NEG} & \texttt{O} \\ 
        \hline
        Unified & \texttt{O} & \texttt{B-POS} & \texttt{I-POS} & \texttt{E-POS} & \texttt{O} & \texttt{O} & \texttt{O} & \texttt{O} & \texttt{O} & \texttt{O} & \texttt{O} & \texttt{S-NEG} & \texttt{O} \\ \Xhline{2\arrayrulewidth}
    \end{tabular}
    \caption{Tagging schemes used in the integrated approaches. ``Joint'' and ``Unified'' refers to joint and unified approaches respectively.}
    \label{tab:tagging}
\end{table*}

Although the importance of solving the complete TBSA task remains significant, existing studies are relatively less and their findings~\cite{D13-1171,D15-1073}, to some extent, discouraged other researchers to do further explorations. However, we think that research efforts should be paid to explore a more integrated model for solving this task, because its two sub-tasks are highly coupled together and the potential of a more integrated model is promising.

In this paper, we investigate the complete task of TBSA and design a novel unified framework to handle it in an end-to-end fashion. The proposed framework involves two stacked Recurrent Neural Networks (RNN). The upper one produces the final tagging results of the TBSA task based on the unified tagging scheme. The lower one performs an auxiliary prediction of target boundaries with the aim for guiding and providing the information to the upper RNN. 
Such design is based on the observation that under the unified tagging scheme, the span information is exactly identical to that under the boundary tagging scheme. Refer to the example in Table~\ref{tab:tagging}, if a word is at the beginning of a target mention under the boundary scheme, i.e., having the tag \texttt{B}, it should also be at the beginning under the unified scheme, i.e., having the tag \texttt{B-POS}. In order to explore such inter-scheme tag dependency, we propose to guide the prediction of the upper RNN for the complete TBSA task with the boundary prediction from the auxiliary task, corresponding to the lower RNN. Specifically, we design a component to encode the dependencies into a transition matrix and use the matrix to map the probability distribution of the boundary prediction to the unified tag space of the TBSA task. Then, we determine the proportions of the obtained boundary-based probability scores in the tagging decision and consolidate them with the probability scores from the upper RNN for final predictions.

We also propose to maintain the consistency of the sentiment of individual words within the same target mention based on a simple gate mechanism. The gate mechanism is designed to explicitly consolidate the features of the current word and the previous word. Since both of the gate here and the transition matrix above need to take reliable boundary prediction for performing well, improving the reliability of such prediction in the lower RNN is supposed to be useful for the complete TBSA task. Therefore, we introduce another component to estimate the potential of a word to be a target word.
Note that as defined by the task~\cite{S14-2004,S15-2082,S16-1002}, an opinion target should always co-occur with opinion words, thus, the words close to the opinion words are more likely to be target words and we obtain additional supervision signals for refining boundary information based on this assumption.

In the experiments, our framework outperforms the state-of-the-art methods and the strongest sequence taggers on several benchmark datasets. We conducted detailed ablation studies to quantitatively demonstrate the effectiveness of the designed components. With some case analysis, we show how our framework can handle some difficult cases with the help of the designed components.

\begin{figure*}
    \centering
    \includegraphics[width=1\textwidth]{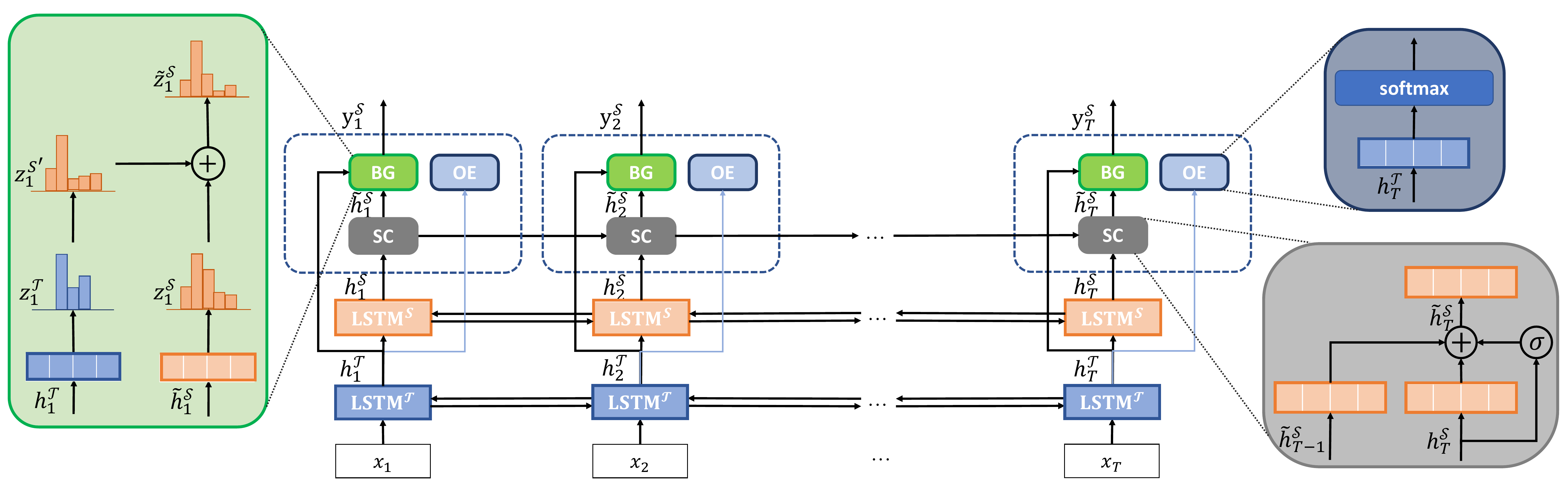}
    \caption{Architecture of the proposed framework.}
    \label{fig:architecture}
\end{figure*}

\section{Our Proposed Framework}
\subsection{Task Definition}

We formulate the complete Target-Based Sentiment Analysis (TBSA) task as a sequence labeling problem, and employ a unified tagging scheme: $\mathcal{Y}^{\mathcal{S}} = \{\texttt{B-POS},\texttt{I-POS},\texttt{E-POS}, \texttt{S-POS}, \texttt{B-NEG},\texttt{I-NEG},\texttt{E-NEG},\\ \texttt{S-NEG}, \texttt{B-NEU},\texttt{I-NEU},\texttt{E-NEU}, \texttt{S-NEU}, \texttt{O}\}$. 
Except \texttt{O}, each tag contains two parts of tagging information: the boundary of target mention, and the target sentiment. For example, $\texttt{B-POS}$ denotes the beginning of a positive target mention, and $\texttt{S-NEG}$ denotes a single-word negative opinion target. 
For a given input sequence $\mathrm{ X}=\{x_1,\dots,x_T\}$ with length $T$, our goal is to predict a tag sequence $\mathrm{ Y}^{\mathcal{S}}=\{y^{\mathcal{S}}_1,\dots,y^{\mathcal{S}}_T\}$, where $y^{\mathcal{S}}_i \in \mathcal{Y}^{\mathcal{S}}$.

\subsection{Model Description}
\subsubsection{Overview}
As shown in Figure~\ref{fig:architecture}, on the top of two stacked RNNs with LSTM cells, our framework designs three tailor-made components, depicted in detail with the callouts, to explore three important intuitions in the task of TBSA. 
Specifically, the upper $\text{LSTM}^{\mathcal{S}}$ is for the complete TBSA task and it predicts the unified tags as output, while the lower $\text{LSTM}^{\mathcal{T}}$ is for the auxiliary task and predicts the boundary tags of target mentions. The boundary prediction from $\text{LSTM}^{\mathcal{T}}$ is used to guide $\text{LSTM}^{\mathcal{S}}$ to make better predictions over the unified tags for the complete task. 


The three key components are named \textbf{B}oundary \textbf{G}uidance (BG) component, \textbf{S}entiment \textbf{C}onsistency (SC) component and \textbf{O}pinion-\textbf{E}nhanced (OE) Target Word Detection component. The BG component takes the advantages of the boundary information provided by the auxiliary task to guide the $\text{LSTM}^{\mathcal{S}}$ for predicting the unified tags more accurately. The SC component is empowered with a gate mechanism to explicitly integrate the features of the previous word into the current prediction, aiming at maintaining the sentiment consistency within a multi-word opinion target. In order to provide boundary information of higher quality, the OE component, following the oberservation that ``opinion targets and opinion words always co-occur'', performs another auxiliary binary classification task to determine if the current word is a target word.

\subsubsection{Target Boundary Guided TBSA}
We employ $\text{LSTM}^{\mathcal{S}}$ with softmax decoding layer for the prediction of the tag sequence. It is observed that the boundary tag can provide important clues for the unified tag prediction. For example, if the current boundary tag is \texttt{B}, denoting the beginning of an opinion target, then the corresponding unified tag can only be $\texttt{B-POS}$, $\texttt{B-NEG}$ or $\texttt{B-NEU}$. Thus, we introduce an additional network $\text{LSTM}^{\mathcal{T}}$ for the target boundary prediction, where the valid tag set $\mathcal{Y}^{\mathcal{T}}$ is \{\texttt{B}, \texttt{I}, \texttt{E}, \texttt{S}, \texttt{O}\}. We link these two LSTM layers so that the hidden representations generated by the $\text{LSTM}^{\mathcal{T}}$ can be directly fed to $\text{LSTM}^{\mathcal{S}}$ as guidance information. 
Specifically, their hidden representations $h^{\mathcal{T}}_t \in \mathbb{R}^{\text{dim}^{\mathcal{T}}_h}$ and $h^{\mathcal{S}}_t \in \mathbb{R}^{\text{dim}^{\mathcal{S}}_h}$ at the $t$-th time step ($t \in [1, T]$) are calculated as follows:
\begin{equation}
\begin{split}
h^{\mathcal{T}}_t &= [\overrightarrow{\text{LSTM}}^{\mathcal{T}}(x_t); \overleftarrow{\text{LSTM}}^{\mathcal{T}}(x_t)],  \\
h^{\mathcal{S}}_t &= [\overrightarrow{\text{LSTM}}^{\mathcal{S}}(h^{\mathcal{T}}_t); \overleftarrow{\text{LSTM}}^{\mathcal{S}}(h^{\mathcal{T}}_t)], t \in [1, T].
\end{split}
\end{equation}
The probability scores $z^{\mathcal{T}}_t \in \mathbb{R}^{|\mathcal{Y}^{\mathcal{T}}|}$ over the boundary tags are calculated by a fully-connected softmax layer:
\begin{equation}
    z^{\mathcal{T}}_t = {\bf p}(y^{\mathcal{T}}_t|x_t) = \mathrm{Softmax}(\mathrm{\bf W}^{\mathcal{T}} h^{\mathcal{T}}_t).
\end{equation}
where the $\text{Softmax}$ denotes the softmax activation function and $\text{\bf W}^{\mathcal{T}}$ is the model parameter. Similarly, the scores over the unified tags $z^{\mathcal{S}}_t \in \mathbb{R}^{|\mathcal{Y}^{\mathcal{S}}|}$ are obtained as below:
\begin{equation}
    z^{\mathcal{S}}_t = {\bf p}(y^{\mathcal{S}}_t|h^{\mathcal{T}}_t) = \mathrm{Softmax}(\mathrm{\bf W}^{\mathcal{S}} h^{\mathcal{S}}_t).
\end{equation}

As mentioned above, the boundary information is supposed to be useful for improving the performance of $\text{LSTM}^{\mathcal{S}}$. \cite{D15-1073} incorporated such boundary information by adding hard boundary constraints in the decoding step of the CRFs model. However, their prediction results are not promising. One reason is that their model employs a hard constraint which is prone to propagating the errors from the tagger of the boundary detection task and thus it decreases the performance of the TBSA tagger. Different from their way of imposing hard constraints, our proposed BG component can absorb the boundary information via boundary guided transition and automatically determine its proportions in the final tagging decision based on the confidence of the target boundary tagger. Firstly, the BG component encodes the constraints into a transition matrix $\text{\bf W}^{tr} \in \mathbb{R}^{|\mathcal{Y}^{\mathcal{T}}| \times |\mathcal{Y}^{\mathcal{S}}|}$.
As we have no prior knowledge about the transition probabilities between the boundary tags and the unified tags, we initially set them equally as follows:
\begin{equation}
\label{eq:position}
\text{\bf W}^{tr}_{i,j}=
\begin{cases} 
      \frac{1}{|\mathcal{B}_i|}, & \text{if } j \in \mathcal{B}_i\\
      0, & \text{Otherwise}
\end{cases}
\end{equation}
where $\mathcal{B}_i$ is the set of valid unified tags coherent with the boundary tag $i$. In this transition matrix, a non-zero element, e.g., $\text{\bf W}^{tr}_{\texttt{B},\texttt{B-POS}}$, denotes the probabilities of the unified tags given the boundary tag, and a zero element, e.g., $\text{\bf W}^{tr}_{\texttt{B},\texttt{I-NEG}}$, suggests that the unified tag cannot be inferred through this transition. After encoding the constraints, the next step is to guide the unified tag prediction with the boundary information. We directly propagate such information to the TBSA tagger by mapping the probability scores of the boundary tag $z^{\mathcal{T}}_t$ to the unified tag space. The transition-based sentiment score $z^{\mathcal{S}^{'}}_t \in \mathbb{R}^{|\mathcal{Y}^{\mathcal{S}}|}$ is obtained as follows:
\begin{equation}
    z^{\mathcal{S}^{'}}_t = (\text{\bf W}^{tr})^{\top} z^{\mathcal{T}}_t
\end{equation}
where the transition operation is equivalent to the linear combination of the row vectors in the transition matrix $\text{\bf W}^{tr}$. Assuming $z^{\mathcal{T}}_t=[1,0,0,0,0]$ (i.e., taking the tag \texttt{B}), the result of the transition is exactly the row vector $\text{\bf W}^{tr}_{\texttt{B},:}$. As the unified tag can be partially derived from the boundary tag, a natural question is how to determine the proportions of the transition-based unified tagging scores $z^{\mathcal{S}^{'}}_t$. Intuitively, if the target boundary score $z^{\mathcal{T}}_t$ is nearly uniform, suggesting that the boundary tagger is not confident to its prediction, the obtained distribution over the unified tags, i.e., $z^{\mathcal{S}^{'}}_t$, will also be close to a uniform distribution and has little meaningful information for the sentiment prediction. To avoid such uninformative boundary transitions, we calculate a proportion score $\alpha_t \in \mathbb{R}$ based on the confidence $c_t$ of the target boundary tagger:
\begin{equation}
\begin{split}
    c_t &= (z^{\mathcal{T}}_t)^{\top} z^{\mathcal{T}}_t \\
    \alpha_t &= \epsilon c_t
\end{split}
\end{equation}
where the hyper-parameter $\epsilon$ denotes the maximum proportions that the boundary-based scores $z^{\mathcal{S}^{'}}_t$ occupy in the tagging decision. Obviously, $c_t$ will be down-weighted if the boundary scores are uniformly distributed. The maximum confidence value is reached if $z^{\mathcal{T}}_t$ is a one-hot vector. The final scores are obtained by combining the boundary-based and model-based unified tagging scores:
\begin{equation}
    \tilde{z}^{\mathcal{S}}_t = \alpha_t z^{\mathcal{S}^{'}}_t + (1-\alpha_t) z^{\mathcal{S}}_t.
\end{equation}

\subsubsection{Maintaining Sentiment Consistency} 
In the traditional target sentiment classification task, the sentiments towards the different words in a given multi-word opinion target are assumed to be identical. However, in the complete TBSA task, such sentiment consistency is not guaranteed since the task is formulated as a sequence tagging/labeling problem. Taking the sentence in Table~\ref{tab:tagging} as an example, there is still some possibility that the word ``Processor'' is labeled with an \texttt{E-NEG} tag due to the independent tagging decisions made by LSTMs. To maintain the sentiment consistency within the same opinion target, we propose to predict the current unified tag using both of the features from the current and the previous time steps. Specifically, we design a Sentiment Consistency (SC) component with a gate mechanism to combine these two feature vectors:
\begin{equation}
\begin{split}
    \tilde{h}^{\mathcal{S}}_t &= g_t \odot h^{\mathcal{S}}_t + (1-g_t) \odot \tilde{h}^{\mathcal{S}}_{t-1} \\
    g_t &= \sigma(\mathrm{\bf W}^g h^{\mathcal{S}}_t + \mathrm{\bf b}^g)
\end{split}
\end{equation}
where $\mathrm{\bf W}^g$ and $\mathrm{\bf b}^g$ are learnable parameters of the SC component, and $\odot$ denotes the element-wise multiplication. $\sigma$ is the sigmoid function. Through the gating, the previous features are considered in the current predictions and such indirect bi-gram dependency can help reduce the probability that the words within the same target hold different sentiments.

\subsubsection{Auxiliary Target Word Detection}
A good boundary tagger for opinion targets is crucial for producing the boundary information of high quality. Here, we introduce the \textbf{OE} component to learn a more robust boundary tagger from another view of the training data. As defined in~\cite{S14-2004,S15-2082,S16-1002}, opinion targets are always collocated with opinion words. Inspired by this, we regard the word as a target word if there is at least one opinion word within the context window of fixed-size $s$ of this word. Then, we train an auxiliary token-level classifier for discriminating target words and non-target words based on the distantly supervised labels and the boundary representations $h^{\mathcal{T}}_t$ are further refined with such supervision signals. The computational process of the \textbf{OE} component is below:      
\begin{equation}
    \begin{split}
        z^{\mathcal{O}}_t &= \text{Softmax}(\text{\bf W}^o h^{\mathcal{T}}_t) \\
        y^{\mathcal{O}}_t &= \arg\max_y z^{\mathcal{O}}_t
    \end{split}
\end{equation}
where $\text{\bf W}^o$ is the model parameter.

\subsection{Model Training}
All the components in our framework are differentiable, thus, the whole framework can be efficiently trained with gradient-based methods. Word/Token-level cross-entropy error is employed as the loss function:
\begin{equation}
    \mathcal{L}^{\mathcal{I}} = -\frac{1}{T} \sum^{T}_{t=1} \mathbb{I}(y^{\mathcal{I},g}_t) \circ \log(z^{\mathcal{I}}_t)
\end{equation}
where $\mathcal{I}$ is the symbol of task indicator and its possible values are $\mathcal{T}$, $\mathcal{S}$, and $\mathcal{O}$. $\mathbb{I}(y)$ represents the one-hot vector with the $y$-th component being 1 and $y^{\mathcal{I},g}_t$ is the gold standard tag for the task $\mathcal{I}$ at the time step $t$. Then, the losses from the main TBSA task and the two auxiliary tasks are aggregated to form the training objective $\mathcal{J}(\theta)$ of the framework:
\begin{equation}
    \mathcal{J}(\theta) = \mathcal{L}^{\mathcal{S}} + \mathcal{L}^{\mathcal{T}} + \mathcal{L}^{\mathcal{O}}.
\end{equation}

\begin{table}[]
    \centering
    \begin{tabular}{cc|c|c|c|c}
    \Xhline{3\arrayrulewidth}
      \multicolumn{2}{c|}{Dataset} & Train & Dev & Test & Total  \\ \hline
        \multirow{3}{*}{$\mathbb{D}_{\mathrm{L}}$} & \# $\texttt{POS}$ & 883 & 104 & 339 & 1326 \\ 
        & \# $\texttt{NEG}$ & 754 & 106 & 130 & 990 \\ 
        & \# $\texttt{NEU}$ & 404 & 46 & 165 & 615 \\ \hline
        \multirow{3}{*}{$\mathbb{D}_{\mathrm{R}}$} & \# $\texttt{POS}$ & 2337 & 270 & 1524 & 4131 \\ 
        & \# $\texttt{NEG}$ & 942 & 93 & 500 & 1535\\ 
        & \# $\texttt{NEU}$ & 614 & 50 & 263 & 927 \\ \hline
        \multirow{3}{*}{$\mathbb{D}_{\mathrm{T}}$} & \# $\texttt{POS}$ & \multicolumn{3}{c|}{-} & 692 \\ 
        & \# $\texttt{NEG}$ & \multicolumn{3}{c|}{-} & 263  \\ 
        & \# $\texttt{NEU}$ & \multicolumn{3}{c|}{-} & 2244  \\ \Xhline{3\arrayrulewidth}
    \end{tabular}
    \caption{Statistics of the datasets.}
    \label{tab:dataset}
\end{table}

\begin{table*}[]
    \centering
    \begin{tabular}{ll|ccc|ccc|ccc}
    \Xhline{3\arrayrulewidth}
        & \multirow{2}{*}{Model} & \multicolumn{3}{c|}{$\mathbb{D}_{\text{L}}$} & \multicolumn{3}{c|}{$\mathbb{D}_{\text{R}}$} & \multicolumn{3}{c}{$\mathbb{D}_{\text{T}}$}\\ \cline{3-11} 
        & & P & R & F1 & P & R & F1 & P & R & F1 \\ \hline \hline
        \multirow{4}{*}{\textbf{Existing Baselines}} & CRF-joint & 57.38 & 35.76 & 44.06 & 60.00 & 48.57 & 53.68 & 43.09 & 24.67 & 31.35 \\
        & CRF-unified & 59.27 & 41.86 & 49.06 & 63.39 & 57.74 & 60.43 & 48.35 & 19.64 & 27.86 \\ 
        & NN-CRF-joint & 55.64 & 34.48 & 45.49 & 61.56 & 50.00 & 55.18 & 44.62 & 35.84 & 39.67 \\
        & NN-CRF-unified & 58.72 & 45.96 & 51.56 & 62.61 & 60.53 & 61.56 & 46.32 & 32.84 & 38.36 \\ \hline
        \multirow{3}{*}{\textbf{Pipeline Baselines}} & CRF-pipeline & 59.69 & 47.54 & 52.93 & 52.28 & 51.01 & 51.64 & 42.97 & 25.21 & 31.73 \\
        & NN-CRF-pipeline & 57.72 & 49.32 & 53.19 & 60.09 & 61.93 & 61.00 & 43.71 & 37.12 & 40.06 \\
        & HAST-TNet & 56.42 & 54.20 & 55.29 & 62.18 & 73.49 & 67.36 & 46.30 & 49.13 & 47.66 \\ \hline
        \multirow{4}{*}{\textbf{Unified Baselines}} & LSTM-unified & 57.91 & 46.21 & 51.40 & 62.80 & 63.49 & 63.14 & 51.45 & 37.62 & 43.41 \\
        &LSTM-CRF-1 & 58.61 & 50.47 & 54.24 & 66.10 & 66.30 & 66.20 & 51.67 & 44.08 & 47.52 \\
        & LSTM-CRF-2 & 58.66 & 51.26 & 54.71 & 61.56 & 67.26 & 64.29 & 53.74 & 42.21 & 47.26 \\
        & LM-LSTM-CRF & 53.31 & 59.4 & 56.19 & 68.46 & 64.43 & 66.38 & 43.52 & 52.01 & 47.35 \\ \hline
        \multirow{5}{*}{\textbf{OURS}}  
        & Base model & 60.00 & 46.85 & 52.61 & 61.48 & 66.16 & 63.73 & 53.02 & 41.47 & 46.50 \\
        & Base model + \textbf{BG} & 58.58 & 50.63 & 54.31 & 67.51 & 66.42 & 66.96 & 52.26 & 43.84 & 47.66 \\
        & Base model + \textbf{BG} + \textbf{SC} & 58.95 & 53.00 & 55.81 & 63.95 & 69.65 & 66.68 & 53.12 & 43.60 & 47.79 \\
        & Base model + \textbf{BG} + \textbf{OE} & 63.43 & 49.53 & 55.62 & 62.85 & 66.77 & 65.22 & 53.10 & 43.50 & 47.78 \\
        & Full model & 61.27 & 54.89 & \textbf{57.90}$^{\natural,\sharp}$ & 68.64 & 71.01 & \textbf{69.80}$^{\natural,\sharp}$ & 53.08 & 43.56 & \textbf{48.01}$^{\sharp}$ \\
        \Xhline{3\arrayrulewidth}
    \end{tabular}
    \caption{Main results of the complete TBSA task. ``Base model'' refers to the stacked LSTMs. The markers $\natural$ and $\sharp$ refer to our full model significantly outperforms \textbf{HAST-TNet} and \textbf{LM-LSTM-CRF} respectively.}
    \label{tab:main_results}
\end{table*}

\section{Experiments}
\subsection{Dataset}
Our model is evaluated on two product review datasets from SemEval ABSA challenges~\cite{S14-2004,S15-2082,S16-1002} and the Twitter dataset. Table~\ref{tab:dataset} gives the statistics of these benchmark datasets. $\mathbb{D}_{\mathrm{L}}$ (SemEval 2014) contains reviews from the laptop domain and the train-test split is the same as the original dataset. $\mathbb{D}_{\mathrm{R}}$ is the union set of the restaurant datasets from SemEval ABSA challenge 2014, 2015 and 2016. The new training dataset is obtained by merging the three years' training datasets and the new testing set is built in the same way. $\mathbb{D}_{\mathrm{T}}$ consists of tweets collected by~\cite{D13-1171}. The ground truth of the opinion target mentions and their sentiments are provided in these datasets. For $\mathbb{D}_{\mathrm{L}}$ and $\mathbb{D}_{\mathrm{R}}$, we regard 10\% randomly held-out training data as the development set. For $\mathbb{D}_{\mathrm{T}}$, we report the ten-fold cross validation results, as done in~\cite{D13-1171,D15-1073}, since there is no standard train-test split for this dataset. 

The gold standard boundary annotations are available for the auxiliary target boundary prediction task. For another auxiliary task, namely, opinion-based target word detection, we employ the existing opinion lexicon\footnote{http://mpqa.cs.pitt.edu/} to provide the opinion words. 

The evaluation metric measures the standard precision (P), recall (R) and F1 score based on the \textbf{exact} match, which means that an output segment is considered to be correct only if it exactly matches with the gold standard span of the target mention and the corresponding sentiment. 

\subsection{Compared Models}
We compare our framework with the following methods:
\begin{itemize}
    \item \textbf{CRF}-\{\textbf{pipeline}, \textbf{joint}, \textbf{unified}\}~\cite{D13-1171}: Conditional Random Fields (CRF) based sequence tagger\footnote{http://www.m-mitchell.com/code/index.html}. ``pipeline'' denotes the pipeline approach. ``joint'' and ``unified'' are the models following the joint tagging scheme and unified tagging scheme respectively.
    \item \textbf{NN-CRF}-\{\textbf{pipeline}, \textbf{joint}, \textbf{unified}\}~\cite{D15-1073}: Enhanced CRF models\footnote{https://github.com/SUTDNLP/NNTargetedSentiment} armed with word embeddings and neural network feature extractors.
    \item \textbf{HAST-TNet}: HAST~\cite{li2018aspect} and TNet~\cite{P18-1087} are the current state-of-the-art models on the tasks of target boundary detection and target sentiment classification respectively. HAST-TNet is the pipline approach of these two models. We use the officially released codes\footnote{Available at: https://github.com/lixin4ever/HAST and  https://github.com/lixin4ever/TNet respectively.} to produce the results.
    \item \textbf{LSTM-unified}: the standard LSTM model adopting the unified tagging scheme.
    \item \textbf{LSTM-CRF-1}~\cite{N16-1030}: LSTM model with CRF decoding layer and no feature engineering is needed. We run the officially released code~\footnote{https://github.com/glample/tagger} and utilize the unified tag set to reproduce the results. 
    \item \textbf{LSTM-CRF-2}~\cite{P16-1101}: LSTM-CRF-2 is similar to LSTM-CRF-1. The difference is that LSTM-CRF-2 employs CNN rather than LSTM to learn the character-level word representations. We run the released code\footnote{https://github.com/XuezheMax/NeuroNLP2} to reproduce the results.
    \item \textbf{LM-LSTM-CRF}~\cite{liu2017empower}: Language model enhanced LSTM-CRF model. It is a competitive model in several sequence tagging tasks. We rerun their code\footnote{https://github.com/LiyuanLucasLiu/LM-LSTM-CRF} and report the tagging results based on the unified tagging scheme.
\end{itemize}

\subsection{Experiment Settings}

\subsubsection{Word Embeddings} 
We use \texttt{GloVe.840B.300d}~\footnote{https://nlp.stanford.edu/projects/glove/} released by~\cite{D14-1162} to initialize the word embeddings, fine-tuned during training. The embeddings of the out-of-vocabulary words are sampled from the uniform distribution $\mathcal{U}$(-0.25, 0.25)~\cite{D14-1181}. 

\subsubsection{Weight Initializations}
The weight matrices in the LSTM units are initialized by following the Glorot Uniform strategy~\cite{glorot2010understanding} and the others are randomly sampled from the uniform distribution $\mathcal{U}$(-0.2, 0.2). Besides, all biases are initialized as 0's.
\subsubsection{Optimization}
Our models are trained up to 50 epochs with Adam~\cite{kingma2014adam}, with $\beta_1 = \beta_2 =
0.9$, and the initial learning rate $\eta_0 = 10^{-3}$. The decay rate is kept the same as the setting in~\cite{N16-1030}. We apply dropout on word embeddings and the ultimate features for prediction. The dropout rates are empirically set as 0.5. The model obtaining the best F1 score on the development set is selected for producing the testing results. 

\subsubsection{Others}
Both of the dimension of the hidden representations $\text{dim}^{\mathcal{T}}_h$ and $\text{dim}^{\mathcal{S}}_h$ are 50. The maximum proportion $\epsilon$ of the boundary-based scores is 0.5. The size of the context window $s$ in the opinion-based target word detection component is 3. The tuning details of $\epsilon$ and $s$ are given later.

\begin{table*}[t]
    \centering
    \resizebox{1.0\textwidth}{!}
    {%
    \begin{tabular}{@{}L{5.8cm}@{~}|@{~}L{1.8cm}@{~}|@{~}L{2.9cm}@{~}|@{~}L{1.8cm}@{~}|@{~}L{2.9cm}@{~}|@{~}L{1.8cm}@{~}|@{~}L{2.9cm}@{}}
    \Xhline{3\arrayrulewidth}
        \multirow{2}{*}{\textbf{Input}} & \multicolumn{2}{c|@{~}}{Base model} & \multicolumn{2}{c|@{~}}{Base model + \textbf{BG}} & \multicolumn{2}{c}{\textbf{Full model}} \\ \cline{2-7}
        & Target & Complete & Target & Complete & Target & Complete  \\ \hline
        1. And the fact that it comes with an \textbf{[}\textcolor{red}{\textit{i5 processor}}\textbf{]}$_{\texttt{POS}}$ definitely speeds things up & \textit{i5 processor} & [\textit{processor}]$_{\texttt{POS}}$ (${\text{\xmark}}$)& \textit{i5 processor} & [\textit{i5 processor}]$_{\texttt{POS}}$ & \textit{i5 processor} & [\textit{i5 processor}]$_{\texttt{POS}}$ \\ \hline
        2. There were small problems with \textbf{[}\textcolor{red}{\textit{mac office}}\textbf{]}$_{\texttt{NEG}}$ . & \textit{mac office} & [\textit{mac}]$_{\texttt{NEG}}$ (${\text{\xmark}}$) & \textit{mac office} & [\textit{mac office}]$_{\texttt{NEG}}$ & \textit{mac office} & [\textit{mac office}]$_{\texttt{NEG}}$  \\ \hline
        3. The \textbf{[}\textcolor{red}{\textit{teas}}\textbf{]}$_{\texttt{POS}}$ are great and all the \textbf{[}\textcolor{red}{\textit{sweets}}\textbf{]}$_{\texttt{POS}}$ are homemade & \textit{teas}, \textit{sweets} & [\textit{teas}]$_{\texttt{POS}}$, [\textit{sweets}]$_{\texttt{POS}}$ & \textit{teas}, \textit{sweets}, \textit{homemade} (${\text{\xmark}}$) & [\textit{teas}]$_{\texttt{POS}}$, [\textit{sweets}]$_{\texttt{POS}}$, [\textit{homemade}]$_{\texttt{POS}}$ (${\text{\xmark}}$) & \textit{teas}, \textit{sweets} & [\textit{teas}]$_{\texttt{POS}}$, [\textit{sweets}]$_{\texttt{POS}}$ \\ \hline
        4. I love the \textbf{[}\textcolor{red}{\textit{form factor}}\textbf{]}$_{\texttt{POS}}$ & NONE & NONE & NONE & NONE & \textit{form factor} & [\textit{form factor}]$_{\texttt{POS}}$ \\ \hline
        5. I blame the \textbf{[}\textcolor{red}{\textit{Mac OS}}\textbf{]}$_{\texttt{NEG}}$ . & \textit{Mac OS} & [\textit{Mac}$_\texttt{NEG}$ \textit{OS}$_{\texttt{NEU}}$] (${\text{\xmark}}$) & \textit{Mac OS} & [\textit{Mac}$_\texttt{NEG}$ \textit{OS}$_{\texttt{POS}}$] (${\text{\xmark}}$) & \textit{Mac OS} & [\textit{Mac OS}]$_{\texttt{NEG}}$ \\ \hline
        6. Also, I personally wasn't a fan of the \textbf{[}\textcolor{red}{\textit{portobello and asparagus mole}}\textbf{]}$_{\texttt{NEG}}$ . & \textit{portobello and asparagus mole} & [\textit{portobello}$_{\texttt{NEG}}$ \textit{and}$_{\texttt{NEG}}$ \textit{asparagus}$_{\texttt{NEG}}$ \textit{mole}$_{\texttt{NEU}}$] (${\text{\xmark}}$) & \textit{portobello and asparagus mole} & [\textit{portobello}$_{\texttt{NEG}}$ \textit{and}$_{\texttt{NEG}}$ \textit{asparagus}$_{\texttt{NEU}}$ \textit{mole}$_{\texttt{NEU}}$] (${\text{\xmark}}$) & \textit{portobello and asparagus mole} & [\textit{portobello and asparagus mole}]$_{\texttt{NEG}}$ \\
        \Xhline{3\arrayrulewidth}
    \end{tabular}}
    \caption{Case analysis. The ``Target'' column contains the results from the auxiliary task of target boundary detection. The ``Complete'' column presents the output of the complete TBSA task, but note that we only show the sentiment part of the unified labels (i.e., \texttt{POS}, \texttt{NEG}, and \texttt{NEU}) and use brackets to indicate the boundary. The marker ${\text{\xmark}}$ denotes the incorrect prediction. }
    \label{tab:case_study}
\end{table*}

\subsection{Results and Analysis}
\subsubsection{Main Results}
Table~\ref{tab:main_results} presents our comparisons with other methods for the complete TBSA task. To
make the comparison fair, we use \texttt{GloVe.840B.300d} as the pre-trained word embeddings for all the baselines requiring word embedding input on all of the datasets. Besides, we align the train/dev/test configurations for all methods. The experimental results suggest that our proposed framework consistently gives the best F1 score across all datasets and significantly outperforms the strongest baselines in most cases.

Compared to HAST-TNet, the pipeline of two state-of-the-art models, our proposed framework achieves 2.6\%, 2.4\% and 0.40\% absolute gains on $\mathbb{D}_{\mathrm{L}}$, $\mathbb{D}_{\mathrm{R}}$ and $\mathbb{D}_{\mathrm{T}}$ respectively, suggesting that a carefully-designed integrated model can be more effective than the pipeline approaches on the TBSA task. 
Three competitive unified sequence taggers (see the third block in Table~\ref{tab:main_results}) are also introduced into the comparative study. Again, our framework outperforms the best of them by 1.7\%, 3.4\% and 0.5\% on the benchmark datasets. We notice that the improvement of our framework on the Twitter dataset is marginal in contrast with the unified baselines. The small gap is reasonable since these models employ additional component (e.g., LSTM or CNN) to learn the character-level word representations, whose capability for representing out-of-vocabulary words has been verified in~\cite{santos2014learning,kim2016character}, while our framework only utilizes the word-level features provided by the pre-trained word embeddings. Similar observation is captured in the comparison with HAST-TNet. We attribute this to the superior modeling power of the CNN applied in TNet when processing the ungrammatical sentences such as tweets and micro-blogs, as pointed out in~\cite{P18-1087}. 

We also notice that the performances of the CRF-based models, especially the recall (R) scores, are quite poor. Armed with the pre-trained word embeddings and neural network feature extractor, the models are slightly improved but the scores are still not promising. 

\subsubsection{Effectiveness of the Proposed Components}
To investigate the effectiveness of the designed components, we conduct ablation study on the proposed framework and the results are listed in the last block of the Table~\ref{tab:main_results}. 
Let us start the discussion from the base model, namely, the stacked LSTMs. We find that the base model always gives superior performance compared to the LSTM-unified. 
This result indicates that the boundary information predicted by the auxiliary LSTM indeed increases the F1 score of the complete TBSA task. 
With the help of the BG component, the performances are improved more significantly and the way we impose the boundary constraints proves effective for yielding more true positives. 
Another interesting finding is that introducing the component \textbf{SC} or \textbf{OE} individually into the ``Base model + \textbf{BG}'' does not bring in too much gains on F1 measure and even hurts the prediction performance on $\mathbb{D}_{\text{R}}$. But putting them together, i.e.,  the ``Full model'', leads to the new state-of-the-art result. 
This result illustrates the necessity of both of the \textbf{SC} and \textbf{OE} components in the boundary guided TBSA. Considering the ``Base model + \textbf{BG} + \textbf{SC}'', the quality of the boundary information may not be accurate without the clues from the \textbf{OE} component, and thus, the \textbf{SC} component tends to incorrectly align the sentiments of both the target words and non-target words. 
For the ``Base model + \textbf{BG} + \textbf{OE}'', the quality of the boundary information obtained from the $\text{LSTM}^{\mathcal{T}}$ is improved but the sentiments of the words within the same target are not fully consistent compared to the ``Full model'' armed with \textbf{SC} component. 
In summary, the \textbf{SC} component and the \textbf{OE} component are complementary to some extent when they are added into the boundary-guided ``Base model + \textbf{BG}''.

\subsubsection{Case Analysis}
Table~\ref{tab:case_study} gives some prediction examples of the base model (i.e., the stacked LSTMs) and the models empowered with our proposed components. As observed in the first input and the second input, the ``Base model'' correctly predicts the target boundary but it fails to produce the right target sentiments, suggesting that linking the two LSTMs for the target boundary prediction and the TBSA task is still insufficient for exploiting the boundary information to improve the performance of the complete TBSA. The ``Base model+\textbf{BG}'' and the ``Full model'', where the boundary constraints are properly imposed via our BG component, can correctly handle these two cases. Although the boundary information can guide the model to predict the sentiment more accurately, there is the possibility that only using the BG component (i.e., ``Base model+\textbf{BG}'') inherits the errors from the lower boundary detection task, e.g., the third and the fourth input. Thus, the boundary information of high quality is crucial for improving the upper TBSA task and our OE component can serve as a simple but effective solution. Besides, we find that maintaining sentiment consistency within the same target mention, especially for whose with several words (e.g., ``\textit{portobello and asparagus mole}'' in the last input), is difficult for the ``Base model'' and ``Base model+\textbf{BG}'', while our ``Full model'' alleviates this issue by employing the SC component to make predictions based on the features from the current and the previous time step. 

\subsubsection{Impact of $\epsilon$ and $s$}
\begin{figure}[t]
    \centering
    \includegraphics[width=0.48\textwidth]{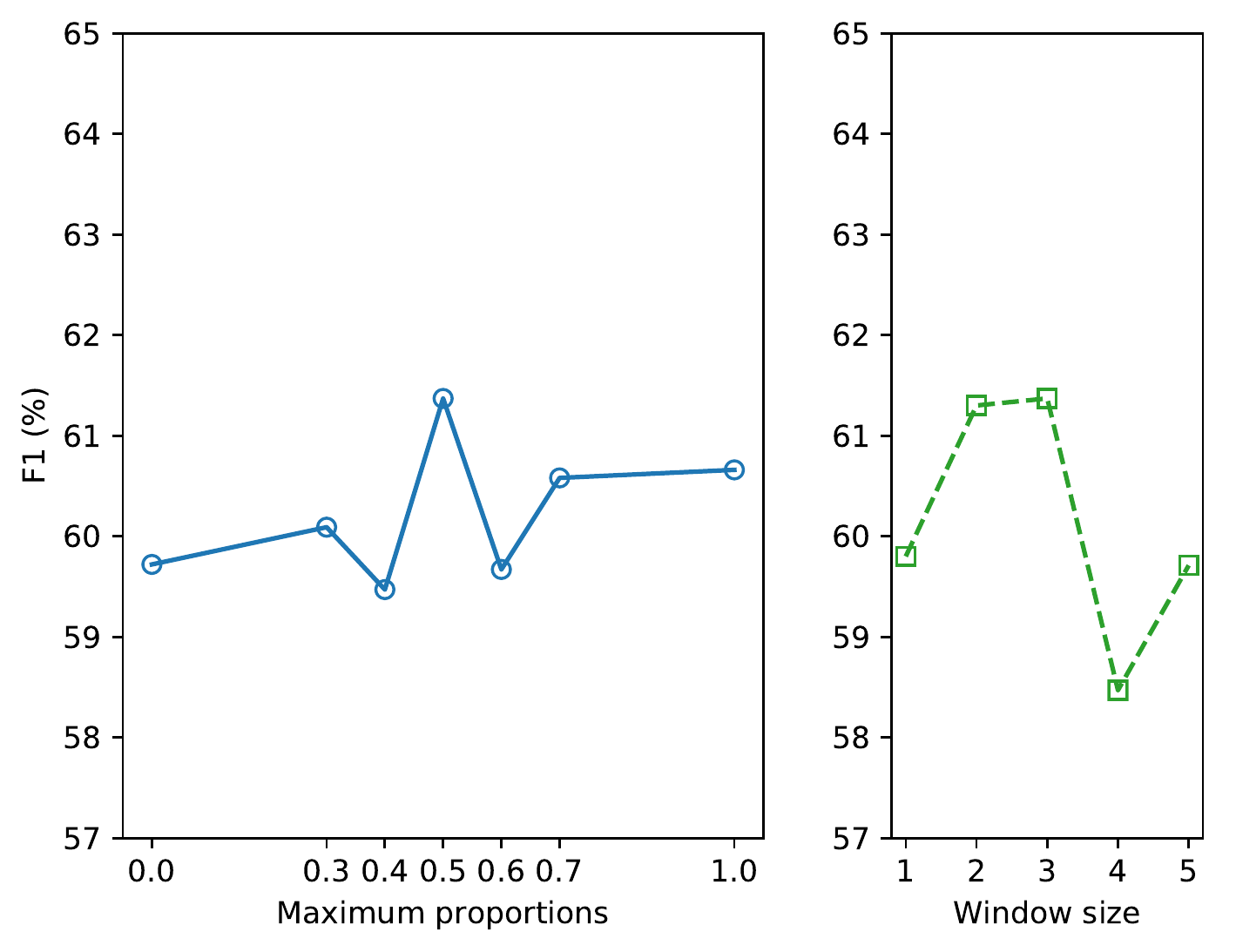}
    \caption{F1 scores (\%) on the development set of $\mathbb{D}_{\mathrm{R}}$ with different $\epsilon$ and $s$ values.}
    \label{fig:tunning}
\end{figure}

Here, we investigate the impacts of the maximum proportion $\epsilon$ of the boundary-based scores and the window size $s$ on the prediction performance. 
Specifically, the experiments are conducted on the development set of $\mathbb{D}_{\mathrm{R}}$, the largest benchmark dataset. 
We vary $\epsilon$ from 0.3 to 0.7, increased by 0.1, and two extreme values 0.0 and 1.0 are also included. The range of the window size $s$ is 1 to 5. 
According to the results given in Figure~\ref{fig:tunning}, we observe that the best results are obtained at $\epsilon$=0.5. The $\epsilon$ value basically affects the importance of the sentiment scores from the BG component in the final tagging decision and 0.5 is a good trade-off between absorbing boundary information and eliminating noises. We also observe that a moderate value of $s$ (i.e., $s$ = 3) is the best for the TBSA task, probably because too large $s$ may enforce the model to attend the larger context and increase the possibility of associating with irrelevant opinion words, on the other hand, too small $s$ is likely not sufficient to involve the potential opinion words.

\section{Related Works}
As mentioned in Introduction, Target-based Sentiment Analysis are usually divided into two sub-tasks, namely, the Opinion Target Extraction task (OTE) and the Target Sentiment Classification (TSC) task. Although these two sub-tasks are treated as separate tasks and solved individually in most cases, for more practical applications, they should be solved in one framework. Given an input sentence, the output of a method should contain not only the extracted opinion targets, but also the sentiment predictions towards them. Some previous works attempted to discover the relationship between these two sub-tasks and gave a more integrated solution for solving the complete TBSA task. Concretely, \cite{D13-1171} employed Conditional Random Fields (CRF) together with hand-crafted linguistic features to detect the boundary of the target mention and predict the sentiment polarity. \cite{D15-1073} further improved the performance of the CRF based method by introducing a fully connected layer to consolidate the linguistic features and word embeddings. However, they found that a pipeline method can beat both of the model with joint training and the unified model. In this paper, we reexamine the task, and proposed a new unified solution which outperforms all previous reported methods. 

\section{Conclusions}
We investigate the complete task of Target-Based Sentiment Analysis (TBSA), which is formulated as a sequence tagging problem with a unified tagging scheme in this paper. The basic architecture of our framework involves two stacked LSTMs for performing the auxiliary target boundary detection and the complete TBSA task respectively. On top of the base model, we designed two components to take the advantage of the target boundary information from the auxiliary task and maintain the sentiment consistency of the words within the same target. To ensure the quality of the boundary information, we employ an auxiliary opinion-based target word detection component to refine the predicted target boundaries. Experimental results and case studies well illustrate the effectiveness of our proposed framework, and a new state-of-the-art result of this task is achieved. We publicly release our implementation at \url{https://github.com/lixin4ever/E2E-TBSA}.

\bibliographystyle{aaai}
\bibliography{aaai19}
\end{document}